\pdfoutput=1
\documentclass[10pt,twocolumn,letterpaper]{article}

\usepackage{cvpr}   
\usepackage[accsupp]{axessibility}

\usepackage{graphicx}
\usepackage[font=normalsize]{subcaption}
\usepackage{amsmath}
\usepackage{amssymb}
\usepackage{booktabs}
\usepackage{algorithm, algorithmic}

\usepackage{array,mathtools, multirow}
\makeatletter
\patchcmd{\ALG@step}{\addtocounter{ALG@line}{1}}{\refstepcounter{ALG@line}}{}{}
\newcommand{\ALG@lineautorefname}{Line}
\makeatother

\newcommand*{\affaddr}[1]{#1} 
\newcommand*{\affmark}[1][*]{\textsuperscript{#1}}
\newcommand*{\email}[1]{\small{\texttt{#1}}}

\usepackage{amsmath,amsfonts,bm}

\def\Figref#1{Figure~\ref{#1}}

\def\eqref#1{equation~\ref{#1}}

\def\1{\bm{1}}

\def\rmX{{\mathbf{X}}}
\def\rmY{{\mathbf{Y}}}

\def\vx{{\bm{x}}}
\def\vy{{\bm{y}}}

\def\mW{{\bm{W}}}

\DeclareMathAlphabet{\mathsfit}{\encodingdefault}{\sfdefault}{m}{sl}
\SetMathAlphabet{\mathsfit}{bold}{\encodingdefault}{\sfdefault}{bx}{n}
\newcommand{\tens}[1]{\bm{\mathsfit{#1}}}
\def\tA{{\tens{A}}}

\newcommand{\Ls}{\mathcal{L}}
\newcommand{\R}{\mathbb{R}}

\DeclareMathOperator*{\argmin}{arg\,min}

\usepackage{xcolor}
\usepackage{makecell}
\usepackage[pagebackref,breaklinks,colorlinks]{hyperref}

\usepackage[capitalize]{cleveref}
\crefname{section}{Sec.}{Secs.}
\Crefname{section}{Section}{Sections}
\Crefname{table}{Table}{Tables}
\crefname{table}{Tab.}{Tabs.}

\def\cvprPaperID{6027} 
\def\confName{CVPR}
\def\confYear{2022}

\begin{document}

\title{ResSFL: A Resistance Transfer Framework for Defending Model Inversion Attack in Split Federated Learning}


\author{%
Jingtao Li\affmark[1], Adnan Siraj Rakin\affmark[1], Xing Chen\affmark[1], Zhezhi He\affmark[2], Deliang Fan\affmark[1], Chaitali Chakrabarti\affmark[1]\\
\affaddr{\affmark[1] School of Electrical Computer and Energy Engineering, Arizona State University, Tempe, AZ}\\
\affaddr{\affmark[2] Department of Computer Science and Engineering, Shanghai Jiao Tong University, Shanghai}
\\
\email{\affmark[1]\{jingtao1, asrakin, xchen382, dfan, chaitali\}@asu.edu};\; \email{\affmark[2]\{zhezhi.he\}@sjtu.edu.cn}
}

\maketitle

\begin{abstract}
This work aims to tackle Model Inversion~(MI) attack on Split Federated Learning~(SFL). SFL is a recent distributed training scheme where multiple clients send intermediate activations~(i.e., feature map), instead of raw data, to a central server.
While such a scheme helps reduce the computational load at the client end, it opens itself to reconstruction of raw data from intermediate activation by the server. Existing works on protecting SFL only consider inference and do not handle attacks during training.
So we propose ResSFL, a Split Federated Learning Framework that is designed to be MI-resistant during training. It is based on deriving a resistant feature extractor via attacker-aware training, and using this extractor to initialize the client-side model prior to standard SFL training. Such a method helps in reducing the computational complexity due to use of strong inversion model in client-side adversarial training as well as vulnerability of attacks launched in early training epochs.
On CIFAR-100 dataset, our proposed framework successfully mitigates MI attack on a VGG-11 model with a high reconstruction Mean-Square-Error of 0.050 compared to 0.005 obtained by the baseline system. The framework achieves 67.5\% accuracy (only 1\% accuracy drop) with very low computation overhead.  Code is released at: \url{https://github.com/zlijingtao/ResSFL}.
\end{abstract}


\section{Introduction}
Collaborative training schemes have become popular in applications where preserving data privacy is very important.  A representative example is federated learning  \cite{mcmahan2017communication}, which has been used in a broad range of computer vision tasks involving private data, such as in human face recognition. 
Split Federated Learning (SFL)\cite{thapa2020splitfed} is a recent collaborative training scheme that combines the merits of Split Learning (SL)\cite{gupta2018distributed} and Federated Learning~(FL). It has significant advantages on computation reduction and memory usage compared to FL \cite{mcmahan2017communication, he2020group} and significantly faster compared to the original SL scheme. 

\begin{figure}[t]
    \centering
    \begin{subfigure}[b]{0.5\textwidth}
         \centering
         \includegraphics[clip,width=0.95\columnwidth]{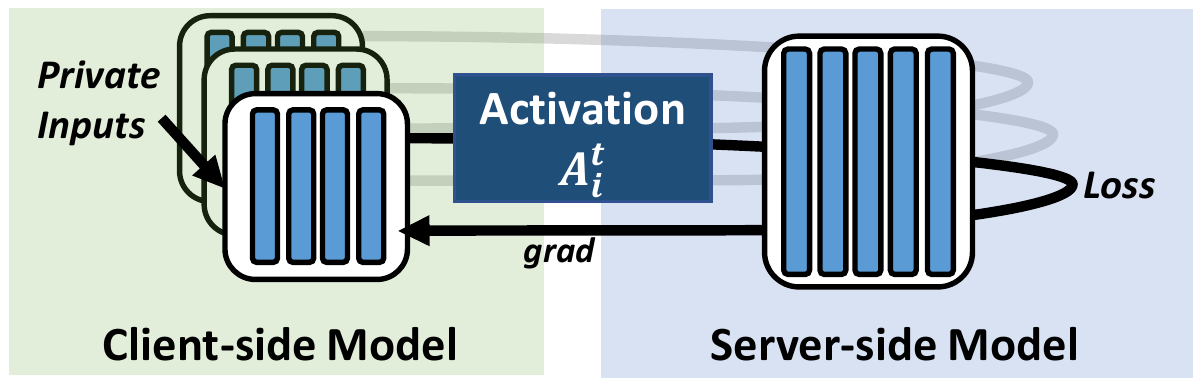}
         \caption{Training process of SFL}
     \end{subfigure}
    \begin{subfigure}[b]{0.5\textwidth}
         \centering
        \vspace{2pt} \includegraphics[clip,width=0.95\columnwidth]{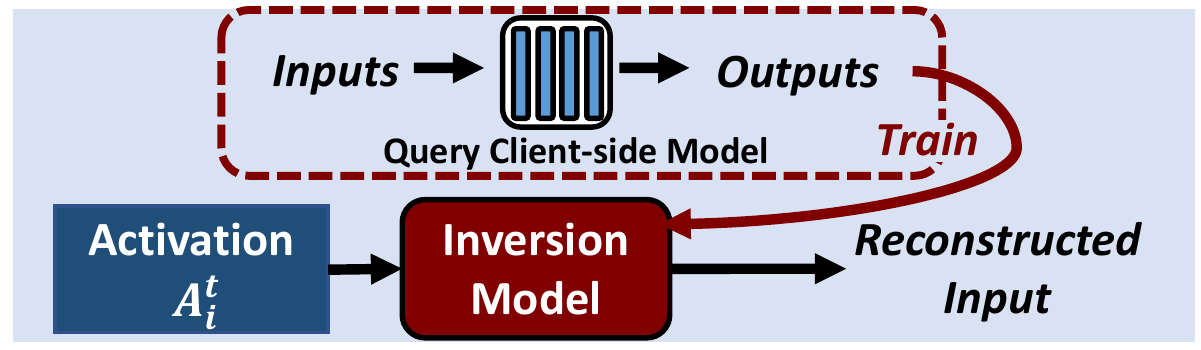}
         \caption{Server perform MI attack}
     \end{subfigure}
    \begin{subfigure}[b]{0.5\textwidth}
         \centering
        \vspace{2pt} \includegraphics[clip,width=0.95\columnwidth]{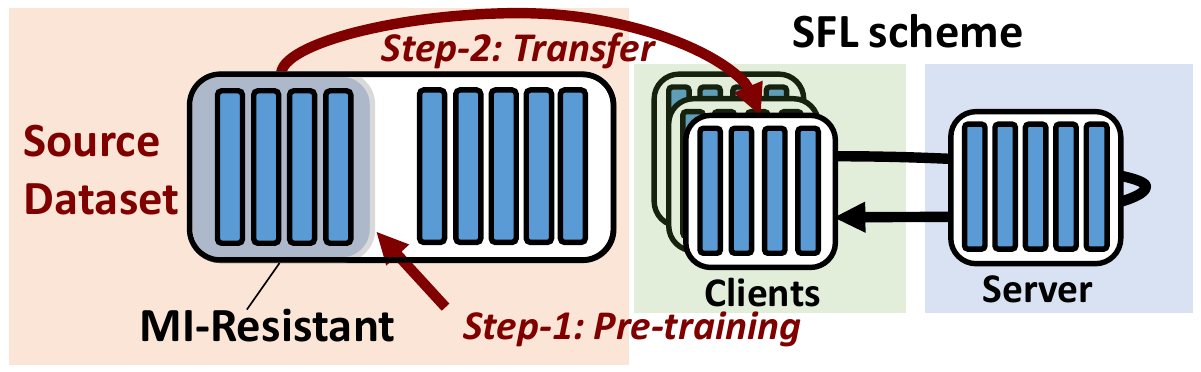}
         \caption{Our proposed ResSFL two-step framework}
     \end{subfigure}
    \caption{ResSFL overview. (a) Multi-client SFL training process. (b) MI attack performed by server. The client-side model is queried to build inversion model and the clients' intermediate activations are used to reconstruct private data. (c) High-level view of the ResSFL framework, consisting of pre-training step and resistance transfer step.}
    \label{fig:SL_scheme}
\end{figure}

In SFL, a neural network model is split into a client-side model and a server-side model. Multiple clients operate on their private inputs using their client-side model and pass the intermediate activations to the server. The server then computes on the more expensive server-side model and passes gradients back to clients. After an epoch, local copies of client-side model are averaged similar to FL. Such a scheme is described in~\Figref{fig:SL_scheme}(a). Thus SFL avoids clients' raw data being sent to the server and also reduces client's computation overhead.


Unfortunately, SFL is reported to be vulnerable to Model Inversion~(MI) attack\cite{he2020attacking, vepakomma2020nopeek}.
Here the honest-but-curious~\cite{paverd2014modelling} server behaves as the attacker. As shown in~\Figref{fig:SL_scheme}(b), by training an inversion model of the client-side model, the server can reconstruct clients' raw data from intermediate activations it received during SFL training.

Prior works provide MI resistance for SFL inference by protecting intermediate activations \cite{vepakomma2019reducing, he2020attacking, vepakomma2020nopeek, titcombe2021practical} or confidence score~(intermediate activations of last softmax layer) \cite{yang2020defending, wang2020improving, wen2021defending}. However, MI resistance at training time is significantly more difficult. While inference-time defense only needs to make the {\it final} model resistant to MI attack, training-time defense must be resistant to MI attack {\it anytime} during the SFL training process. This is because the honest-but-curious server has access to all intermediate activations
and can launch an attack anytime.

To defend against MI attack during training and achieve low computational capability of the 
clients, we present ResSFL, a two-step Split Learning framework that is resistant to MI attacks. The first step uses attacker-aware training to develop MI-resistant feature extractor and  the second step uses this extractor to initialize client-side training prior to standard SFL-based training. An overview of our proposed framework is depicted in~\Figref{fig:SL_scheme}(c).

The  attacker-aware training emulates a strong attacker using a strong inversion model and adds bottleneck layers \cite{eshratifar2019bottlenet} to the inversion model to reduce the feature space.
To reduce the computational complexity of such a process at the client-side and address vulnerability issues in early training epochs,  the attacker-aware training scheme is implemented on an expert device and the MI resistant model is used to initialize the client side model. The clients then implement a lite-version attacker-aware training to achieve good accuracy on the new task while maintaining the resistance of the expert. 
Our contributions can be summarized as follows:
\begin{itemize}
    \item \textbf{Two-step resistant SFL framework.} We present a novel two-step resistant SFL framework based on the proposed attacker-aware training and resistance transfer. To the best of our knowledge, this is the {first work} that successfully mitigates \textbf{training-time MI attack} while achieving good accuracy in SFL.

    \item\textbf{Attacker-aware Training Method.} We use a combination of (i) strong inversion model  to mimic the MI attack behavior and (ii) bottleneck layers to shrink the large feature space resulting in good MI resistance and accuracy.
    
    \item \textbf{Protect Training by Resistance Transfer.} We use transfer learning to protect early-epoch vulnerability of attacker-aware training as well as reduce high computation cost of client-end training with complex inversion models.  
    

\end{itemize}

\section{Background}
\subsection{Split Learning}
Split Learning\cite{gupta2018distributed} is a  collaborative training scheme that is suitable for training using low-end edge devices. By splitting the neural network model into a client-side model and a server-side model, most of the computations can be offloaded to the server. We use cut-layer to denote the split location. \Figref{fig:SL_scheme} (a) shows a SL scheme with cut-layer of 4, implying that the client-side model has 4 layers.

In the state-of-the-art variant of SL called \textit{Split Federated Learning} (SFL)~\cite{thapa2020splitfed}, multiple clients participate in the training process in parallel. The clients'  models are averaged at the beginning of each epoch similar to FL. The parallelism makes it much faster than the original scheme \cite{gupta2018distributed}, where only one client can train at a time and passes the updated client-side model to the next in a round-robin fashion. We specifically consider the \textit{SFL-V2} scheme in \cite{thapa2020splitfed} and refer to it as SFL in the rest of the paper.

\subsection{Threat Model}
\label{section: threat_model}

In the SFL scheme, each client has a local copy of the client-side model denoted as $C^i$ for client $i$; the server-side model is denoted as $S$. The training process executes for a total of $T$ epochs with client models being synchronized by  averaging the local copies at the beginning of each epoch at the server; the synchronized central copy is denoted as $C^*$. The intermediate activations of all training data from client $i$ at epoch $t$ ($t>0$) is denoted as $\tA^i_t$.

The server uses MI attack \cite{fredrikson2015model} to reconstruct the raw inputs. We only consider the training-based MI attack~\cite{fredrikson2015model, yang2019neural} where an auxiliary dataset with similar data distribution is required. The other optimization-based MI attack \cite{dosovitskiy2016inverting, he2020attacking} has weaker performance and is included only in the supplementary material. To perform MI attack, at epoch $t$, the attacker uses an inversion model $G$ that maps the feature space of $\tA^i_t$ to input space. Model $G$ is trained to minimize Mean-Square-Error (MSE) loss between the ground-truth image and the reconstructed image. We assume the server use the validation dataset as the auxiliary dataset, $\vx_{aux}$. The optimization process of $G$ is:

\begin{equation}
    G = \argmin_{G} MSE(G(C(\vx_{aux})), \vx_{aux})
    \label{eq:decoding_MI attack}
\end{equation}

Then, the trained inversion model $G$ is used to reconstruct the intermediate activation $\tA^i_t$ sent by client $i$ to reveal its private training data. 

\begin{equation}
    \vx^*_{priv} = G(\tA^i_t)
    \label{eq:MI attack_recon}
\end{equation}

\noindent \textbf{Assumption I. HBC  condition.} We assume the server is honest-but-curious (HBC) \cite{paverd2014modelling}, that is,  it follows the protocol properly and sends back computationally correct results. However, it records all the intermediate activation $\tA^i_t$ sent by the client $i$. We rule out malicious conditions in \cite{pasquini2021unleashing} since a malicious server can neutralize any possible defense.

\noindent \textbf{Assumption II. White-box assumption.} The server has full query access and \textit{white-box} assumption on client-side model $C^*$,  the training algorithm including defensive methods, architecture information and model parameters.
The assumption is practical since the server can also act as a participating client with its own private data.

\noindent \textbf{Assumption III. Increased Strength of Inversion Model.} We assume that the server can try different model architectures to instantiate the inversion model for MI attack. Unlike previous works \cite{yang2019neural, vepakomma2020nopeek, titcombe2021practical} which use a single architecture (usually very simple), we allow the server to use inversion models with increased complexity, as shown in Table~\ref{tab:decoder_tier}. Note that since the server has no dearth of computational resources, it can easily support high complexity models.

\begin{table}[htbp]
\caption{Description of inversion models with increasing complexity L0 $\rightarrow$ L3.}

\label{tab:decoder_tier}
\vspace{-5pt}
\begin{center}
\begin{small}
\resizebox{1.0\linewidth}{!}{
\begin{tabular}{cccc} 
 \toprule
 \textbf{} & \textbf{Depth} & \textbf{Width} & \textbf{FLOPs\textsuperscript{*}}\\
 \midrule
 L0 &2$\times$ Conv2D + Conv2DTranspose &16 Channels &3.6M\\
 \midrule
 L1 &2$\times$ BasicBlocks + Conv2DTranspose &16 Channels &5.1M\\
 \midrule
 L2 &4$\times$ BasicBlocks + Conv2DTranspose &32 Channels &18.7M\\
 \midrule
 L3 &6$\times$ BasicBlocks + Conv2DTranspose &64 Channels &76.2M\\
 \bottomrule
\end{tabular}
}
\end{small}
\end{center}
\vskip 0.02in
\small \textsuperscript{*}: FLOPs are measured using an input of size [1,128,8,8], which corresponds to intermediate activation of VGG-11 model with cut-layer of 2. In comparison, the client-side model has 21.5 M FLOPs. Architecture details are provided in supplementary material.

\vskip -0.1in
\end{table}

\subsection{Resistance Definition}
Similar to previous works on MI attack \cite{yang2019neural, wen2021defending}, we define \textit{MI resistance} as the MSE between the ground-truth image $x_{priv}$ and the reconstructed image $x^*_{priv}$ generated using inversion model $G$. We choose the inversion model  from Table~\ref{tab:decoder_tier} that achieves the best performance. Empirically, we set an MSE value of 0.02 as the resistance target.




\section{Motivation}
\label{sec:motivation}
\subsection{High Computation Cost}
Attacker-aware training is a popular defense method adopted in defending adversarial attacks~\cite{goodfellow2014explaining, he2019parametric}. It is also used for protecting confidence score from MI attack during inference \cite{yang2020defending}. Here an inversion model is used to train the ``purifier'' that modifies the  confidence score to fool the MI attack.

While this method can be used during SFL training, an inversion model needs access to ground-truth data, meaning the inversion model must be kept locally at client-side. Simulating a weak attacker using the low end L0 or L1 inversion models during training is fine but does not help defend against a strong attacker (as shown later in \Figref{fig:adv_expert}). On the other hand, simulating a strong attacker has a high computation cost. For example, using the L3 decoder costs 76.2M FLOPs, which is over 3x higher than the client-side model (21.5M FLOPs). Such a method is clearly not acceptable for SFL, which is designed for resource-constrained clients.

\subsection{Early-Epoch Vulnerability}


We note that early epochs of the training process can be vulnerable even with a proper defense. 
Evaluation with our proposed attacker-aware training 
shows that while the system is able to build up enough resistance at the end of the training process, the early stages show very low resistance (shown later in \Figref{fig:baseline_early}). This is because resistance building is a slow progressive process, similar to  accuracy.
Thus during the early training epochs, the model has vulnerable interfaces that can be exploited for an MI attack. 





\section{Defense Framework}

\begin{figure*}[!ht]
    \vspace{-2pt}
    \centering
    \includegraphics[width=1.0\linewidth]{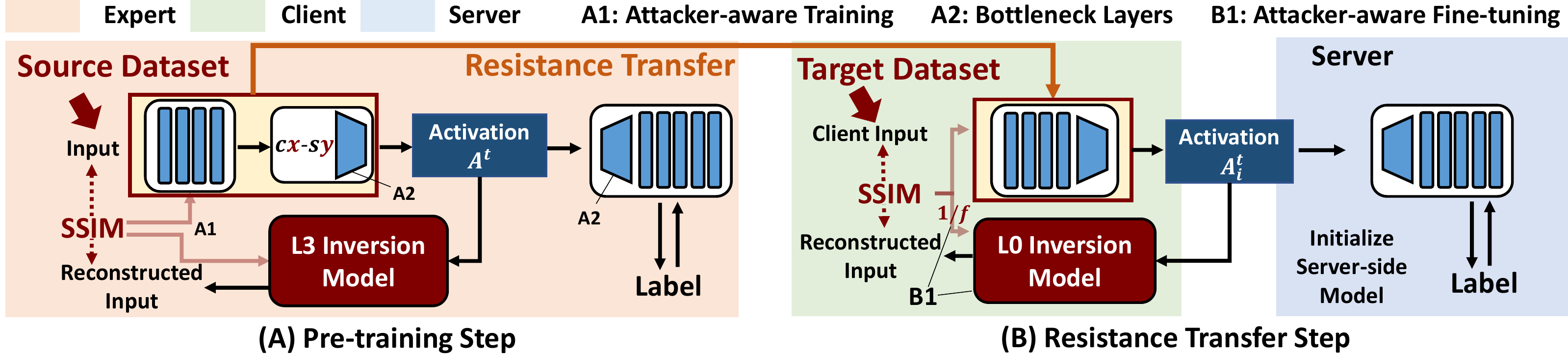}
    \caption{The proposed ResSFL scheme. The expert builds a MI-resistant client-side model using attacker-aware training in step (A). This is followed by a  resistance transfer step (B) that solves a new task by reusing the client-side model obtained in (A). Note: (A1) consists of two training steps in Algorithm~\ref{alg:adv_train}, and (B1) is described in section~\ref{sec:transfer}.}
    \label{fig:AdvSL_scheme}
\end{figure*}

To address the high computational cost of implementing strong inversion models at the client side and early-epoch vulnerability issues, we build a two-step framework called ResSFL. 
It consists of a pre-training step, where we build up a feature extractor with strong MI resistance, and a follow-up resistance transfer step, where the resistant feature extractor is used to initialize the client-side model of the SFL scheme. \Figref{fig:AdvSL_scheme} describes the proposed scheme. Specifically, an expert (i.e. one powerful client or a third party) that has a high computation budget handles the pre-training step on a source task (task 1). The pre-training step utilizes attacker-aware training (A1) and bottleneck layers (A2). 
Then, the resistant feature extractor is used as an initialization for SFL scheme on a new task (task 2). A low complexity attacker-aware fine-tuning method (B1) is used to update the client-side model to adapt to the new task with high accuracy while preserving MI resistance.

\subsection{Attacker-aware Training}\label{sec:adv_train}


The proposed  attacker-aware training includes an inversion model $D^i$ parameterized by $\mW_D^i$ for client $i$.  $D^i$ takes the intermediate activation $\tA^i$ as input and  generates the reconstructed image $\vx^*_i$ -- similar to what a real attacker would do. The attacker-aware training with  $D^i$ takes the Min-Max optimization form of:

\begin{equation}
\begin{aligned}
\min_{\mW_{C}^i,\mW_S} \max_{\mW_D^i} &  \underbrace{\sum_{i=1}^{N}{\Ls_{CE}(S(\mW_{S};C^i(\mW_{C}^i;\vx_i), \vy_i)}}_{\text{Cross-Entropy Loss}} +\\
& \lambda \times \underbrace{ \sum_{i=1}^{N}{\R(D^i(\mW_D^i;C^i(\mW_C^i;\vx_i)), \vx_i)}}_{\text{Inversion Score}}\\
\end{aligned}
\label{eq:objective}
\end{equation}
where $\mW_C^i,\mW_S$ denote the parameters for client-side and server-side models. $\Ls_{\vx_i, \vy_i}$ is cross-entropy loss calculated on client $i$'s private data ($\vx_i, \vy_i$). $\R$ is the score function which evaluates the quality of the reconstructed images compared to ground-truth images $\vx_i$. $\lambda$ is a constant that controls the strength of the inversion score.

From equation~\ref{eq:objective}, we see that computing the inversion score requires access to the private image $\vx_i$, meaning the second optimization must be done locally at client side in a SFL scheme.  We use structural similarity index (SSIM) score \cite{zhao2016loss} for the $\R$ function to evaluate the quality of the image reconstruction. SSIM has a better correlation with human perceived image quality compared to MSE and is widely adopted in image restoration tasks. 

\begin{algorithm}
\caption{Attacker-aware Training} \label{alg:adv_train}
\begin{algorithmic}[1]
\REQUIRE $\quad$ For $N$ clients (N = 1 in pre-training step), instantiate private training data ($\rmX_i, \rmY_i$) for $1, 2, ..., N$, client-side model $C^{i}$, local inversion model $D^{i}$ and server-side model $S$. $\lambda$ is the strength of the inversion score.

\FUNCTION{\textsc{Train} ($\rmX_i$, $\rmY_i$)}{
\STATE initialize $C^{i}, D^{i}, S$
\FOR{epoch $t\gets 1$ to num\_epochs}{
    \STATE $C^{*} = \frac{1}{N}\sum_{i=1}^{N}{C^{i}}$ 
    \FOR{client $i\gets 1$ to $N$ \textbf{in Parallel}}{
    \STATE $C^{i} \gets C^{*} $ 
    \FOR{step $s\gets 1$ to num\_batches}{
     

    \STATE Image batch ($\vx_i, \vy_i$) $\gets$ ($\rmX_i, \rmY_i$)
    \STATE $\tA^i_{t} = C^{i}(\vx_i)$
    \STATE $score= \R(D^i(\mW_D^i;\tA^i_{t}), \vx_i)$    
    \STATE $\max_{\mW_D^i}(score)$  
    \STATE $loss = \Ls_{CE}(S(\mW_{S};\tA^i_{ t}), \vy_i)$
    
    \STATE $\min_{\mW_{C}^i,\mW_S}(loss + \lambda score)$  
    }
    \ENDFOR
    
    }
\ENDFOR
}
\ENDFOR
}
\ENDFUNCTION

\end{algorithmic}
\end{algorithm}

Algorithm~\ref{alg:adv_train} shows the detailed process of attacker-aware training applied on a SFL scheme. $N$ clients compute in parallel (line 7), and client-side model synchronization (line 4) is done at the beginning of each epoch.
We adopt the same strategy as in~\cite{goodfellow2014generative} to solve the Min-Max optimization problem. Each client $i$ undergoes two training steps in each epoch $t$ (corresponding to line 11 and line 13). 
The first training step (line 11) adapts the inversion model $D^i$ by maximizing the score $\R$ by keeping $C^i$ fixed.
The second training step (line 13) is the standard training process, where $W_C^i$ and $W_S$ are updated to minimize the cross-entropy loss. Here, we add the inversion model's score as a regularization term. 
We limit the regularization effect to $W_C^i$ by keeping $D^i$ fixed, similar to \cite{goodfellow2014generative}, which fixes the discriminator when training the generator. The strength of the regularization is controlled by $\lambda$.


\noindent
{\bf Addition of Bottleneck Layers:} 
The accuracy and resistance performance of the attacker-aware training outlined above is not as good, as shown later in \Figref{fig:adv_expert}.
The large feature space of the intermediate activation makes it more vulnerable to MI attack. In fact, it has been shown in \cite{mo2020querying} that attacking intermediate activation rather than confidence score, can result in better MI attack performance.

So our approach is to reduce the dimension of intermediate activations through the use of \textit{bottleneck layers} introduced in \cite{eshratifar2019bottlenet}. We implement bottleneck layers using a pair of Conv2D layers $l_{in}$ and $l_{out}$. $l_{in}$ has input channel size same as original channel size of the intermediate activation, and output channel size of $C$, while $l_{out}$ has input channel size of $C$ and output channel size same as the input channel size of $l_{in}$.  Both layers have kernel size of 3.
For simplicity, we use $Cx$-$Sy$ to denote  bottleneck layers with channel size of $x$ and stride of $y$.

While use of bottleneck layer in mitigating MI attack is new, use of only bottleneck layers does not improve performance, as much as will be shown later in Table~\ref{tab:comparison}. However a combination of  attacker-aware training and bottleneck layers  gives us way better performance in terms of both resistance and accuracy. Thus, we present attacker-aware training with bottleneck layers as the method on which to build up model resistance.

\subsection{Attacker-aware fine-tuning}
\label{sec:transfer}
Typically in transfer learning,  a complex task is transferred to simple tasks and so domain transfer can be done by  freezing the client-side model and fine-tuning the server-side model. However, freezing the client-side model can cause significant accuracy drop. Another option is to allow a small learning rate, which leads to a compromised resistance performance.
To balance accuracy and resistance, we allow the initial model to be tunable and propose a cheaper version of attacker-aware training with low learning rate called \textit{attacker-aware fine-tuning}, to preserve the resistance.
We only allow a low learning rate of 0.005 (compared to a learning rate of 0.02 for rest of the model) on the client-side model and apply the attacker-aware training using the least complex L0 inversion model. Also, we reduce the frequency of inversion model training (line 11 in Algorithm~\ref{alg:adv_train}) to once every 5 training steps ($f=5$) to reduce the overhead of training inversion model at the client-side.

\section{Experiments}

\noindent\textbf{Settings.} To simulate a multi-client SFL, we randomly partition the training dataset evenly to create training datasets (private data) for each client. The server uses validation dataset to validate the model as well as perform the MI attack. We use Stochastic Gradient Descent (SGD) for client-side model and server-side model, and set learning rate to 0.05 with proper learning rate decay. We use Adam optimizer\cite{kingma2014adam} with learning rate of 0.001 for training both the inversion model in performing MI attack as well as in attacker-aware training. Each model is trained for 200 epochs.
Our scheme has two hyperparameters: inversion score strength parameter $\lambda$ for attacker-aware training and $Cx$-$Sy$ for the bottleneck layers. 
We refer to {\it baseline} methods as those that apply attacker-aware training from scratch in SFL, instead of the two-step ResSFL method.  We fix the model architecture to VGG-11 and cut-layer setting to 2 in this section.

\subsection{Build MI-resistant Model (Pre-training)}

\begin{figure}[!ht]
    \vspace{-2pt}
    \centering
    \includegraphics[width=1.0\linewidth]{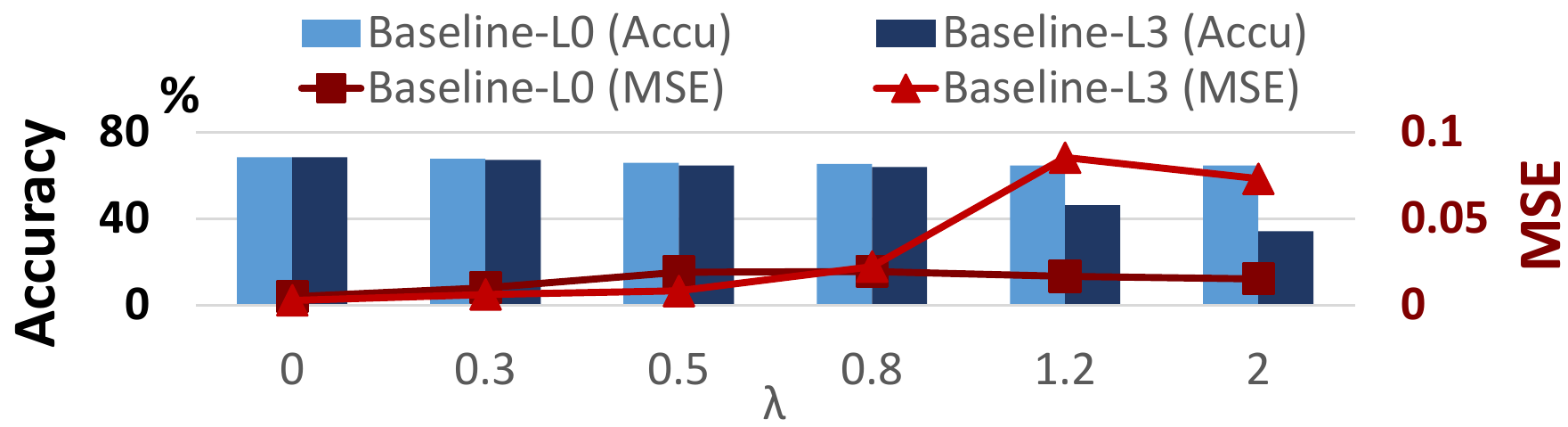}
    \caption{Baseline attacker-aware training methods with L0 (weak) and L3 (strong) inversion models without adding bottleneck layers, on VGG-11 with cut-layer of 2 on CIFAR-100 datasets.}
    \label{fig:baseline_overfit}
\end{figure}

\noindent\textbf{Inversion Model Complexity.} \Figref{fig:baseline_overfit} shows that using a weak inversion model in baseline attacker-aware training (without bottlneck layers) suffers from low resistance (low MSE). We found that the inversion score of the weak inversion model is stuck at a very low value implying a bad reconstruction which translates to poor inversion ability. 
In contrast, using a strong inversion model (L3) closely simulates the attacker and provides high MI resistance for a large $\lambda$. As shown in \Figref{fig:baseline_overfit}, the resistance improves significantly for inversion model L3 when $\lambda = 1.2$ but with significant drop in accuracy.

\begin{figure}[!ht]
    \vspace{-2pt}
    \centering
    \begin{subfigure}[b]{0.5\textwidth}
         \centering
         \includegraphics[clip,width=0.9\columnwidth]{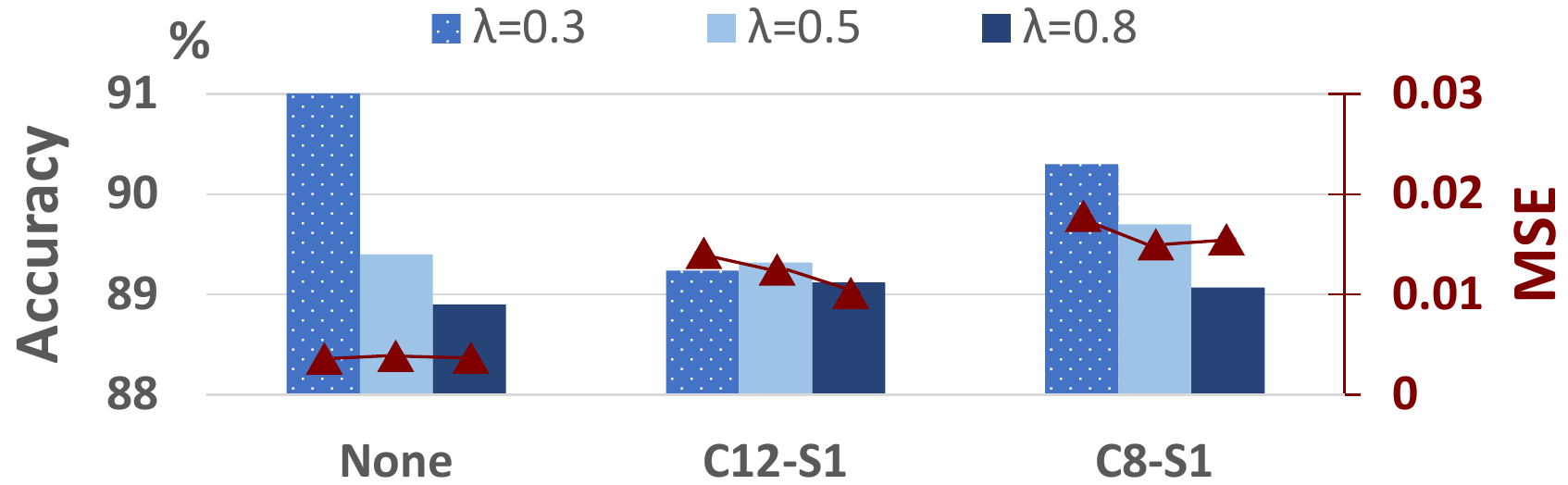}
         \caption{CIFAR-10}
         \vspace{10pt}
     \end{subfigure}
    
    \begin{subfigure}[b]{0.5\textwidth}
         \centering
         \includegraphics[clip,width=0.9\columnwidth]{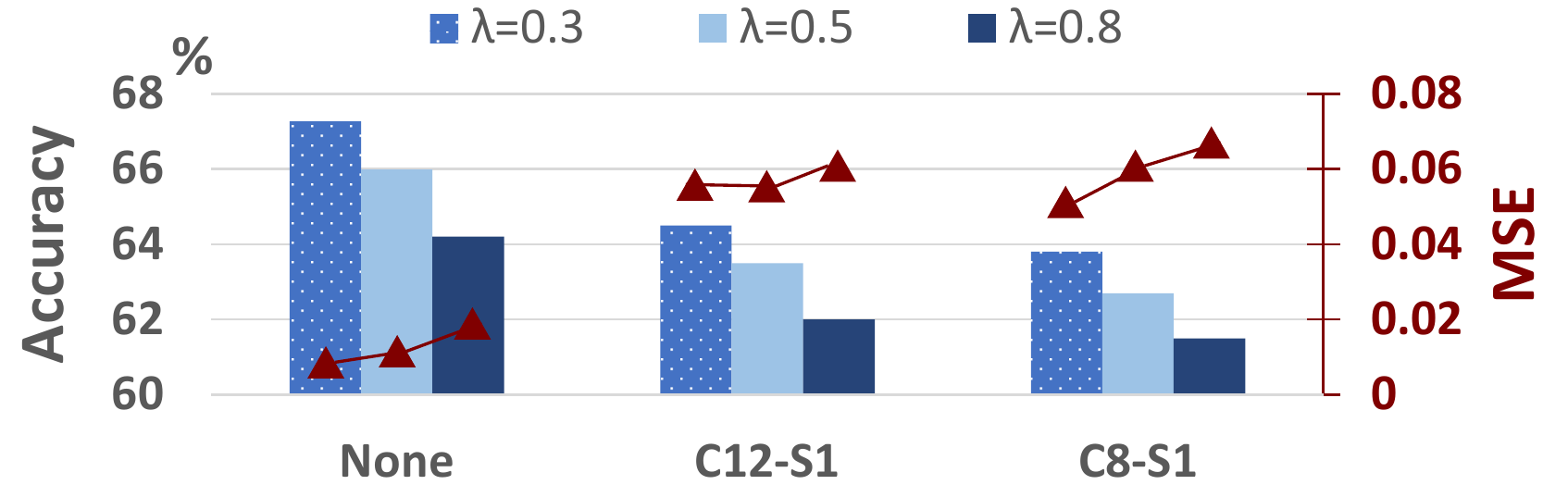}
         \caption{CIFAR-100}
     \end{subfigure}
    \caption{Attacker-aware training applied on VGG-11 with cut-layer of 2 for CIFAR-10/CIFAR-100 datasets. $Cx-Sy$ stands for channel size $x$ and stride of $y$.}
    \label{fig:adv_expert}
\end{figure}

\noindent\textbf{Adding Bottleneck Layers.} For the pre-training step, we apply the proposed attacker-aware training together with bottleneck layers, and train a VGG-11 model with cut-layer of 2.
We vary the bottleneck setting and regularization strength $\lambda$ in equation~\ref{eq:objective} and show corresponding results in \Figref{fig:adv_expert}. Clearly, bottleneck settings ($C12$-$S1$ and $C8$-$S1$) achieves much better resistance than without Bottleneck (None). On CIFAR-10 \& CIFAR-100, attacker-aware training with a narrow bottleneck setting of $C8$-$S1$ can increase the MSE to above 0.02 while still maintaining over 90\% accuracy.

\subsection{Transfer MI resistance (SFL training)}

\noindent\textbf{Different Transfer Strategies.}
We transfer the pre-trained resistant model from two source tasks CIFAR-10 and CIFAR-100 to different target tasks CIFAR-100/CIFAR-10\cite{krizhevsky2009learning}, Facescrub\cite{ng2014data}, SVHN\cite{netzer2011reading} and MNIST\cite{lecun1998mnist} on a 2-client SFL scheme. For FaceScrub and MNIST, we scale the original image to 32x32x3 so that it can be fed into the same client-side model that was pre-trained on CIFAR-10/CIFAR-100.
The performance of resistance transfer is shown in Table~\ref{tab:vgg_transferlearn}. 
The proposed attacker-aware fine-tuning achieves a good balance between accuracy and resistance compared to simple strategies such as \textit{freeze} and \textit{simple fine-tuning} with small learning rate (set at 0.005). For instance, on CIFAR-100, compared to simple fine-tuning,  our proposed attacker-aware fine-tuning achieves only 0.4\% lower accuracy with 0.028 higher MSE. And on SVHN, our proposed method achieves almost no drop in accuracy with only 0.004 lower MSE than {\it freeze}.
 
\noindent\textbf{Transfer Performance to Different Datasets.}
The resistance transfer performance is excellent when transferring from CIFAR-10 to CIFAR-100.
For example, CIFAR10 to CIFAR-100 transfer achieves a very high MSE of 0.050 with only 1\% accuracy drop. This is because CIFAR-10 is more difficult to achieve resistance while CIFAR-100 is easier as shown in \Figref{fig:adv_expert}. So a model being resistant on CIFAR-10 should still be resistant on CIFAR-100. Same for the SVHN dataset, where we successfully reach MSE of 0.045 with almost no accuracy drop. 
However, resistance transfer fails when transferring a simple task to a hard task. For example in CIFAR-100 to CIFAR-10, the MSE is very low at 0.006 on CIFAR-10 even if we only {\it freeze} the client-side model. 
It also shows that the resistance performance weakens when transferring from a source dataset with multiple input channel to a target dataset with a single input channel (i.e. MNIST).
We plan to investigate this in the near future.

\noindent\textbf{Generalization to Multiple Clients.}
We generalize the experiments from 2 clients to more client cases. As shown in Table~\ref{tab:multi_client_transfer}, when the number of clients increase to 20, the proposed scheme achieves similar resistance performance with slight drop in accuracy. Using a client sampling technique, we can extend to 100 clients.


\begin{table}[htbp]
\caption{Resistance transfer results for multiple clients from CIFAR-10 to SVHN and CIFAR-100 dataset of VGG-cut2 model.}
\label{tab:multi_client_transfer}
\vspace{-5pt}
\begin{center}
\begin{small}
\resizebox{0.65\linewidth}{!}{
\begin{tabular}{ccccc} 
 \toprule
 \multirow{2}{*}{\textbf{N}}& \multicolumn{2}{c}{\textbf{SVHN}}& \multicolumn{2}{c}{\textbf{CIFAR-100}}\\
 \cmidrule{2-3}\cmidrule{4-5}
 & \textbf{Accu} & \textbf{MSE} & \textbf{Accu} & \textbf{MSE}\\
 \midrule
  2 &96.0  &0.045  &67.5 &0.050 \\
  5 &95.8  &0.041  &66.9 &0.050 \\
  10 &95.5  &0.041  &66.6 &0.050 \\
  20 &95.5  &0.041  &66.6 &0.048 \\
  100$^a$&95.4  &0.040  &65.8 &0.046 \\
 \bottomrule
\end{tabular}
}
\end{small}

\end{center}
\vskip -0.1in
\centerline{\footnotesize{$^a$ with a sampling rate of 10\% per round.}}
\end{table}


\begin{table*}[htbp]
\caption{Transfer resistance of pre-trained VGG-11 model with cut-layer of 2 on a two-client SFL scheme. Results of fine-tuning client-side model with lite-version attacker-aware training are shown in the last three columns.}
\label{tab:vgg_transferlearn}
\vspace{-5pt}
\begin{center}
\begin{small}
\resizebox{0.75\linewidth}{!}{
\begin{tabular}{cccrlrlrl} 
 \toprule
 \multirow{2}{*}{\textbf{Source Task}}& \multirow{2}{*}{\textbf{Target Task}}& \multirow{2}{*}{\makecell{\textbf{Baseline} \\\textbf{Accuracy}}}& \multicolumn{2}{c}{\textbf{Freeze}}&\multicolumn{2}{c}{\makecell{\textbf{Simple} \\\textbf{Fine-tuning}}}& \multicolumn{2}{c}{\makecell{\textbf{Attacker-aware} \\\textbf{Fine-tuning}}}\\
 \cmidrule{4-9}
 & &  & \textbf{Accu} & \textbf{MSE} & \textbf{Accu} & \textbf{MSE} & \textbf{Accu} & \textbf{MSE}\\
 \midrule
 CIFAR-10& SVHN & 96.1 &92.2 &0.049 &95.9 &0.037 &96.0 &0.045 \\
 CIFAR-10& CIFAR-100 & 68.5 &62.4 &0.077 &67.9 &0.022 &67.5 &0.050 \\
 CIFAR-10& FaceScrub & 82.2 &63.5  &0.021  &68.1  &0.015 &65.2 &0.021 \\

  CIFAR-10& MNIST & 99.6 &99.6  &0.003 &99.6 &0.002 &99.4 &0.003 \\
  \midrule
   CIFAR-100& SVHN & 96.1 &93.6 &0.050 &96.0  &0.023 &95.8 &0.039 \\
  CIFAR-100& FaceScrub & 82.2 &65.0 &0.019 &69.1  &0.015 &67.8  &0.021  \\
 CIFAR-100& CIFAR-10 & 91.9 &89.5 &0.006 &91.5 &0.004  &90.8 &0.005 \\
 CIFAR-100& MNIST & 99.6 &99.6  &0.003  &99.6 &0.002  &99.4 &0.003 \\

 \bottomrule
\end{tabular}
}
\end{small}
\end{center}
\vskip -0.1in
\end{table*}

\subsection{Training-time Resistance Performance}
\begin{figure}[!ht]
    \vspace{-2pt}
    \centering
    \includegraphics[width=1.0\linewidth]{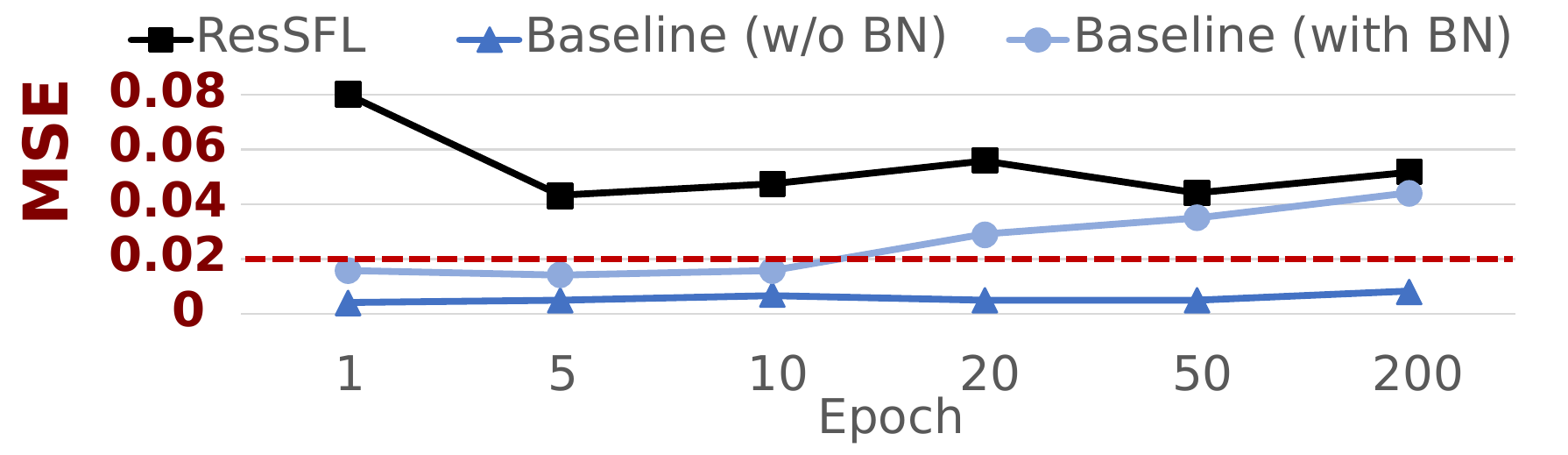}
    \caption{Resistance performance of ResSFL and two baseline methods (with and without bottleneck layers) against training-time MI attack as a function of epoch number.}
    \label{fig:baseline_early}
\end{figure}

To showcase how our framework based on resistance transfer addresses the training-time MI attack, we perform the MI attack at different epochs during SFL training. We use the VGG-cut2 model as the representative model for our ResSFL framework.
For comparison, we use two versions of the baseline method with $\lambda=0.3$ with and without $C8$-$S1$ bottleneck layers. 
\Figref{fig:baseline_early} shows the performance of the baseline method and the ResSFL on MI attack at different epochs (t = 1, 5, ..., 200). ResSFL framework has a steady MI resistance during the training-time, and is consistently better than the target resistance of 0.02 MSE (the red dashed line).

\subsection{Evaluation of Existing Methods}

\begin{table*}[htbp]
\caption{Comparison with previous defenses against MI attack on VGG-11 with cut-layer of 2 in a two-client SFL scheme, on CIFAR-100 dataset with original accuracy of 68.5\% (w.o. any defense). All previous methods fail to get a value over 0.02 MSE while keeping the accuracy drop to be less than 5\%. In contrast, our proposed ResSFL achieves 0.050 MSE with only 1\% accuracy drop.}

\label{tab:comparison}
\vspace{-5pt}
\begin{center}
\begin{small}
\resizebox{0.92\linewidth}{!}{
\begin{tabular}{cccccccccc} 
 \toprule
  \multicolumn{2}{c}{} &\textbf{Accuracy} & \textbf{\makecell{MSE \\(L0)}} & \textbf{\makecell{MSE \\(Best)}} &\multicolumn{2}{c}{} &\textbf{Accuracy} & \textbf{\makecell{MSE \\(L0)}} & \textbf{\makecell{MSE \\(Best)}}\\
 \midrule
 
 \multirow{3}{*}{Laplacian\cite{titcombe2021practical}} & $b=0.05$ &65.4 &0.009 &0.006 &\multirow{3}{*}{AdvNoise\cite{wen2021defending}} &$\epsilon=0.05$ &65.8 &0.012 &0.007\\
   &$b=0.08$ &62.2 &0.012 &0.008 & &$\epsilon=0.08$ &64.6 &0.026 &0.012\\
   &$b=0.10$ &58.4 &0.016 &0.011 & &$\epsilon=0.10$ &62.0 &0.037 &0.018\\
 \midrule
 \multirow{3}{*}{Dropout\cite{he2020attacking}} & $p=0.15$ &64.2 &0.010 &0.006 &\multirow{3}{*}{DistCorr\cite{vepakomma2020nopeek}} &$\alpha=1.0$ &63.6 &0.014 &0.009\\
   &$p=0.20$ &61.6 &0.011 &0.007 & &$\alpha=1.5$ &63.2 &0.036 &0.014\\
   &$p=0.25$ &57.8 &0.012 &0.009 & &$\alpha=2.0$ &62.1 &0.051 &0.019\\
 \midrule
 \multirow{3}{*}{TopkPrune\cite{yang2019neural}} & $k=50$ &65.4 &0.007 &0.004 &\multirow{3}{*}{\makecell{Bottleneck \\ Layers\cite{eshratifar2019bottlenet}}} &$c16$-$s1$ &64.4 &0.008 &0.008\\
   &$k=60$ &61.1 &0.007 &0.004 & &$c8$-$s1$ &63.4 &0.020 &0.013\\
   &$k=70$ &50.4 &0.008 &0.005 & &$c4$-$s1$ &58.0 &0.032 &0.020\\
\midrule
   \textbf{ResSFL}&$\lambda$0.3-$c8$-$s1$ &\textbf{67.5} &\textbf{0.072} &\textbf{0.050} & & & & &\\
 
 \bottomrule
\end{tabular}
}
\end{small}
\end{center}
\vskip -0.1in
\end{table*}

Existing methods on protecting MI on SFL target protection of intermediate activations using perturbation-based methods and regularization-based methods.
Perturbation-based methods include Laplacian noise \cite{titcombe2021practical}, dropout \cite{he2020attacking}, topk-prune described in \cite{yang2019neural} and adversarial noise \cite{wen2021defending}. The method in~\cite{titcombe2021practical} adds noise to the intermediate activation, where the noise follows a Laplacian distribution parameterized by scale $b$ (location parameter $\mu$ is kept at 0). \cite{he2020attacking} uses a mask (each element takes 0 with probability $p$ otherwise 1) and multiplies it element-wise with the intermediate activation. Topk-prune preserves top $k$ percent elements while setting others to zero. \cite{wen2021defending} crafts adversarial noise using FGSM\cite{goodfellow2014explaining} on a surrogate inversion model (we use L3 inversion model) and adds it to the intermediate activation, we use the $\epsilon$ to scale the gradient's sign.
Another line of works apply regularization-based method to build model's intrinsic resistance, similar to our proposed attacker-aware training. Works based on information correlation such as \cite{wang2020improving} and \cite{vepakomma2020nopeek} use mutual information and distance correlation to regularize the training. However, \cite{wang2020improving} is not suitable for SFL as it requires a lot of local computation and
so we only include \cite{vepakomma2020nopeek} in Table~\ref{tab:comparison}.

We vary hyper-parameter settings for each method and keep accuracy in a reasonable range. Except for perturbation-based methods that directly apply on a trained model, we retrain the model and test the MI resistance during inference. As shown in  Table~\ref{tab:comparison}, all previous defensive methods cannot achieve more than 0.02 MSE using both L0 and the best inversion model in MI attacks with accuracy drop of less than 5\%. Distance correlation method performs well against simple L0 inversion model, but fails badly under a strong MI attack using the best inversion model.




\begin{table*}[htbp]
\caption{Transfer resistance of pre-trained models with different architectures on a two-client SFL scheme, from source task CIFAR-10 to target task CIFAR-100. We include FLOPs (measured by feeding a single image), and two popular image quality metrics SSIM and Peak Signal-to-Noise Ratio (PSNR) (using the best inversion model); higher means  better image quality. }
\label{tab:vgg_transferlearn_arch}
\vspace{-5pt}
\begin{center}
\begin{small}
\resizebox{0.7\linewidth}{!}{
\begin{tabular}{cccccccc} 
 \toprule
 \textbf{Topology} & \textbf{\makecell{ResSFL \\ Setting}} & \textbf{FLOPs} & \makecell{\textbf{Original} \\\textbf{Accuracy}} & \makecell{\textbf{ResSFL} \\\textbf{Accuracy}}  & \textbf{MSE} & \textbf{SSIM} & \textbf{PSNR}\\
\midrule
 VGG-cut1 &$\lambda$0.5-$c1$-$s1$ &2.1M  &68.5 &52.3 &0.041 &0.553  &14.4 \\
 VGG-cut2 &$\lambda$0.3-$c8$-$s1$ &21.5M  &68.5 &67.5 &0.050 &0.547  &14.2 \\
 VGG-cut3 &$\lambda$0.3-$c12$-$s1$ &41.6M  &68.5 &67.8 &0.038 &0.539 &15.3\\
\midrule
 Res-cut2 &$\lambda$2.0-$c1$-$s2$ &5.3M  & 67.8 &55.2 &0.052 &0.554 &13.6 \\
 Res-cut3 &$\lambda$0.3-$c1$-$s1$&10.2M & 67.8 &62.0  &0.048 &0.556 &13.9 \\
 Res-cut4 &$\lambda$0.5-$c2$-$s1$&15.1M  & 67.8 &65.6  &0.043 &0.719  &15.6 \\
\midrule
 Mob-cut2 &$\lambda$0.5-$c1$-$s1$&3.7M  & 71.4&67.3  &0.020&0.787&17.6\\
 Mob-cut3 &$\lambda$0.5-$c2$-$s1$&18.7M  & 71.4&70.9  &0.037 &0.717 &15.6\\
 Mob-cut4 &$\lambda$0.3-$c4$-$s1$&31.6M  & 71.4&70.8  &0.062&0.538 &13.3\\
 \bottomrule
\end{tabular}
}
\end{small}
\end{center}
\vskip -0.1in
\end{table*}

\subsection{Cost Analysis}
\label{section:discussion}
The number of parameters and floating point operations (FLOPs) of different schemes are listed in Table~\ref{tab:perturb_train}.
The computation overhead of inversion model training depends on its complexity and updating frequency. Using the proposed attacker-aware fine-tuning, ResSFL has much less computation overhead with 27\% FLOPs and 20\% parameters compared to the the original SFL scheme, and use 0.01\% parameters and 0.17\% FLOPs compared to the popular FL scheme \cite{mcmahan2017communication}.

\begin{table}[htbp]
\caption{Cost analysis of competing schemes on VGG-11 with cut-layer 2 on CIFAR-100. }
\label{tab:perturb_train}
\vspace{-5pt}
\begin{center}
\begin{small}
\resizebox{1.0\linewidth}{!}{
\begin{tabular}{lrlrlc} 
 \toprule
 \textbf{Scheme}& \multicolumn{2}{c}{\textbf{\makecell{Parameters}}}& \multicolumn{2}{c}{\textbf{\makecell{FLOPs}}} & \textbf{\makecell{Resistance}}\\
 \midrule
 FL & 9.8M &(1.00x) & 153.7M &(1.00x) & High\\
 SFL & 76.0K &(0.01x) & 20.9M &(0.14x) & Low\\
 \textbf{ResSFL} & \textbf{91.5K} &\textbf{(0.01x)} & \textbf{26.5M} &\textbf{(0.17x)} & \textbf{High}\\

 \bottomrule
\end{tabular}
}
\end{small}
\end{center}
\vskip -0.1in
\end{table}

\section{Ablation Study}


\noindent\textbf{Choice of Topology.}
We extend the ResSFL framework to other topologies (with different architecture and different cut-layer settings). The second column in Table~\ref{tab:vgg_transferlearn_arch} presents the settings on VGG-11 architecture with cut-layer of 1 and 2, ResNet-20 architecture with cut-layer of 2, 3 and 4, and MobileNet-V2 architecture with cut-layer of 2, 3, and 4 are presented. We act as the expert and tune the channel size/stride size of bottleneck and inversion score strength $\lambda$, to train client-side models towards the resistance target of 0.02 MSE on CIFAR-10 dataset. For example, for ResNet20 architecture with cut-layer of 3, the target resistance is achieved with  $C1$-$S1$ bottleneck and $\lambda$=0.3.
The right three columns in Table~\ref{tab:vgg_transferlearn_arch} show the resistance achieved on CIFAR-100 dataset using resistance transfer for different topologies. We notice that for the same architecture, increasing the cut-layer gets a better accuracy, and better resistance most of the time. Across architecture choices, the MobileNet-V2 performs typically well. Applying ResSFL on MobilNet-V2 with cut-layer of 2 helps achieve a very good accuracy as well as a super high MSE with very small number of FLOPs.
VGG-11 architecture can also achieve a very good resistance with only 1\% accuracy drop.
Thus, different topologies have different
computation complexities, accuracy and resistance tradeoffs, and the topology design space needs further investigation.
\section{Conclusion}

\begin{figure}[!ht]
    \vspace{-2pt}
    \centering
    \includegraphics[width=1.0\linewidth]{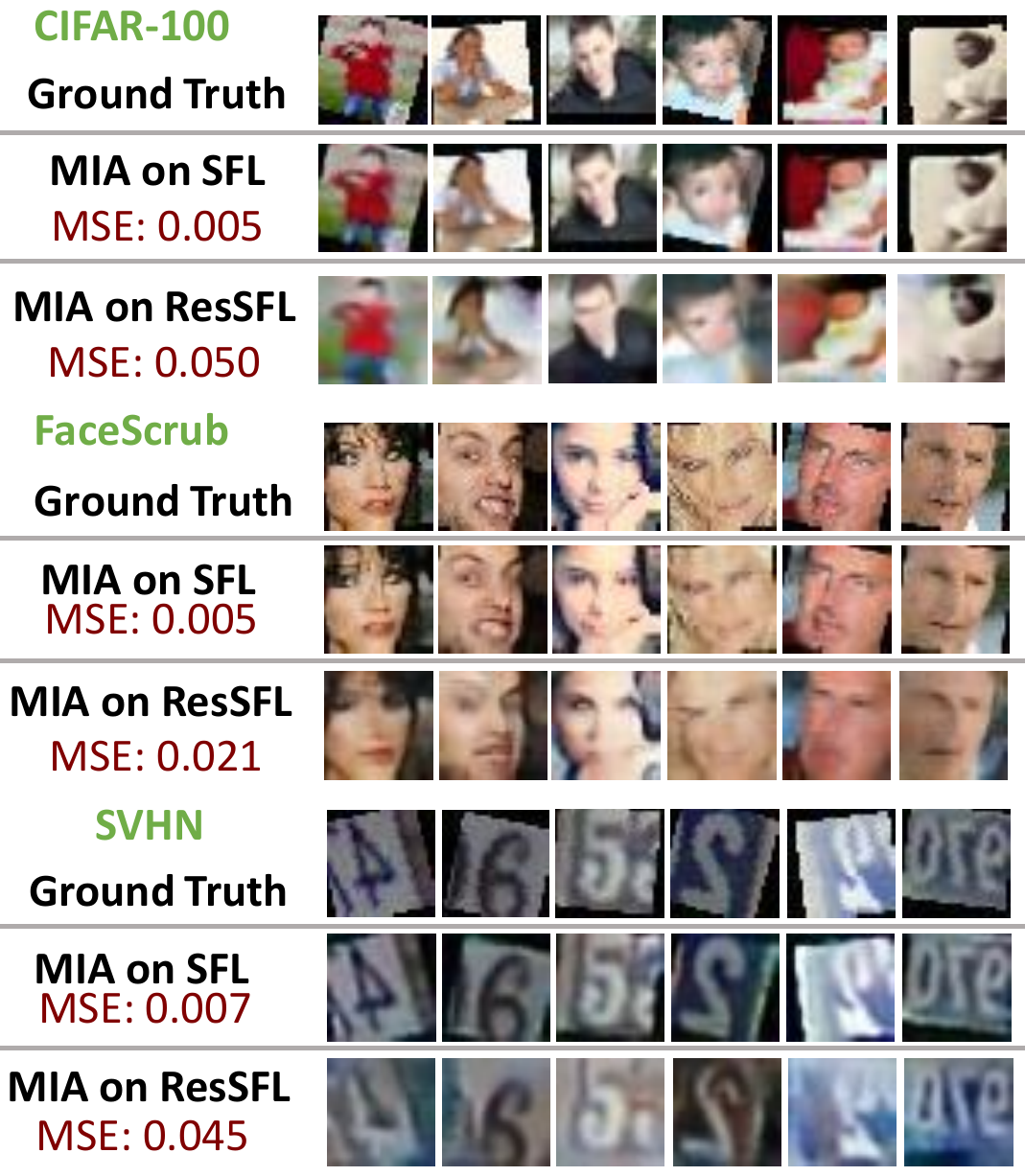}
    \caption{Visualization of MI attack reconstructed images. Model architecture is VGG-11 with cut-layer of 2.}
    \label{fig:teaser_face}
\end{figure}

We present ResSFL, a two-step SFL framework that consists of an attacker-aware training that achieves high MI resistance by an expert and then transferring the resistance to the clients to improve their MI resistance during SFL training.
We show that a combination of attacker-aware training with a complex inversion model and addition of bottleneck layers to the model helps the expert build up strong MI resistance. Transferring this resistance to the clients not only reduces the computational demands at the client-end but also addresses the vulnerability due to attacks in earlier training epochs. We show using the attacker-aware fine-tuning can achieve better balance between accuracy and resistance.
We also show that the framework can support multiple clients and can be generalized to different architecture-cut-layer configurations.
Finally, we show that ResSFL applied to VGG11 model on CIFAR-100 dataset achieves 0.050 MSE compared to 0.005 MSE obtained by the baseline scheme with only 1\% accuracy drop; the corresponding reconstructed faces are very noisy, as shown in \Figref{fig:teaser_face}, and demonstrate the power of the proposed ResSFL scheme.

\newpage
{\small


}

\end{document}